\documentclass[11pt]{article}
\usepackage{coling2018}
\usepackage{times}
\usepackage{url}
\usepackage{latexsym}

\usepackage{graphicx}
\usepackage{comment}
\usepackage{paralist}
\usepackage{footmisc}
\usepackage{wrapfig}
\usepackage{enumitem}
\usepackage{multirow}
\usepackage{xspace} 
\usepackage{fancyvrb} 
\usepackage{caption} 
\usepackage[color=red!5]{todonotes}

\usepackage[skins]{tcolorbox}
\usepackage{xstring}
\usepackage{soul}
\usepackage[normalem]{ulem} 
\usepackage{algpseudocode,algorithm} 
\usepackage{tikz-qtree-compat}
\usetikzlibrary{shapes, fit, arrows.meta}
\usepackage{xcolor,colortbl} 
\usepackage{tabularx}
\usepackage{adjustbox}

\algnewcommand{\LineComment}[1]{\State \(\triangleright\) #1} 
\AtBeginEnvironment{algorithmic}{\scriptsize} 
\algrenewcommand\alglinenumber[1]{\scriptsize #1:} 
\tikzset{>=latex}

\newcommand{\inlinetext}[1]{``\textit{#1}''}
\newcommand{\tuple}[3]{$\langle$#1; #2; #3$\rangle$}
\newcommand{\texttuple}[3]{$\langle$\textit{#1}; \textit{#2}; \textit{#3}$\rangle$}

\newcommand{\conc}{$\mathbin{\|}$}

\newcommand{\ruleSubordinationPost}{Subordinated Clauses with Closing Subordinative Clauses\xspace}

\newenvironment{definitionbox}[1]{\begin{tcolorbox}[title=#1, fonttitle=\bfseries, colback=red!5!white, colframe=red!75!black, colbacktitle=red!75!black]}{\end{tcolorbox}}
\newenvironment{examplebox}[1]{\begin{tcolorbox}[title=#1, fonttitle=\bfseries, colback=blue!5!white, colframe=blue!75!black, colbacktitle=blue!75!black]}{\end{tcolorbox}}

\title{Graphene: Semantically-Linked Propositions\\ in Open Information Extraction}

\author{Matthias Cetto\textsuperscript{1}, Christina Niklaus\textsuperscript{1}, Andr\'{e} Freitas\textsuperscript{2}, \and Siegfried Handschuh\textsuperscript{1} \\
  \textsuperscript{1} Faculty of Computer Science and Mathematics, University of Passau\\
  {\tt \{matthias.cetto, christina.niklaus, siegfried.handschuh\}{\tt @uni-passau.de}}\\
  \textsuperscript{2} School of Computer Science, University of Manchester\\
  {\tt andre.freitas@manchester.ac.uk}
\\}

\date{}

\begin{document}
\maketitle
\begin{abstract}
We present an Open Information Extraction (IE) approach that uses a two-layered transformation stage consisting of a clausal disembedding layer and a phrasal disembedding layer, together with rhetorical relation identification. In that way, we convert sentences that present a complex linguistic structure into simplified, syntactically sound sentences, from which we can extract propositions that are represented in a two-layered hierarchy in the form of \textit{core relational tuples and accompanying contextual information} which are \textit{semantically linked via rhetorical relations}. In a comparative evaluation, we demonstrate that our reference implementation Graphene outperforms state-of-the-art Open IE systems in the construction of correct n-ary predicate-argument structures. Moreover, we show that existing Open IE approaches can benefit from the transformation process of our framework.
\end{abstract}

\section{Introduction}

%
%
\blfootnote{
    %
    %
    %
    
    \hspace{-0.65cm}  
    This work is licenced under a Creative Commons 
    Attribution 4.0 International Licence.
    Licence details:
    \url{http://creativecommons.org/licenses/by/4.0/}
     
    %
}



Information extraction (IE) turns the unstructured information expressed in natural language (NL) text into a structured representation \cite{Jurafsky:2009:SLP:1214993} in the form of relational tuples that consist of a set of arguments and a phrase denoting a semantic relation between them, e.g. \texttuple{Barack Obama}{served as}{the 44th President of the US}. Traditional IE systems have concentrated on identifying and extracting relations of interest by taking as input the target relations, along with hand-crafted extraction patterns or patterns learned from hand-labeled training examples. 
As this manual labor scales linearly with the number of target relations, such a supervised approach does not scale to large, heterogeneous corpora which are likely to contain a variety of unanticipated relations. To tackle this issue, \newcite{Banko07} introduced Open IE as a new extraction paradigm that allows for a domain-independent discovery of relations in large amounts of text by not depending on any relation-specific human input. Instead, detecting the relations is part of the problem. State-of-the-art Open IE systems (see Section \ref{relWork}) identify relationships between entities in a sentence by matching patterns over either shallow syntactic features in terms of part-of-speech (POS) tags and noun phrase (NP) chunks or dependency tree structures. However, particularly long and syntactically complex sentences pose a challenge for current Open IE approaches. By analyzing the output of such systems (see Figure \ref{ComparativeAnalysis}), we observed three common shortcomings.


\begin{figure}[!ht]
  \centering
    \scriptsize
    \begin{BVerbatim}[commandchars=\\\{\},codes={\catcode`$=3\catcode`_=8}]
He nominated Sonia Sotomayor on May 26, 2009 to replace David Souter; she was confirmed on August 6, 2009, 
becoming the first Supreme Court Justice of Hispanic descent. 

OLLIE:
(1) she                was confirmed on                   August 6, 2009
(2) He                 nominated Sonia Sotomayor on       May 26
(3) He                 nominated Sonia Sotomayor          2009
(4) He                 nominated 2009 on                  May 26
(5) Sonia Sotomayor    be nominated 2009 on               May 26
(6) He                 nominated 2009                     Sonia Sotomayor
(7) 2009               be nominated Sonia Sotomayor on    May 26    

ClausIE:
(8)  He     nominated        Sonia Sotomayor on May 26 2009 to replace David Souter 
(9)  she    was confirmed    on August 6 2009 becoming the first Supreme Court Justice of Hispanic descent
(10) she    was confirmed    becoming the first Supreme Court Justice of Hispanic descent

\textbf{Graphene:}
\textbf{(11) #1    0    he    nominated    Sonia Sotomayor}
\textbf{("a)     S:PURPOSE     to replace David Souter }
\textbf{("b)     S:TEMPORAL    on May 26, 2009 }
\textbf{(12) #2    0    she    was confirmed	}
\textbf{("a)     S:TEMPORAL    on August 6, 2009 }
\textbf{(13) #3    0    she    was becoming    the first Supreme Court Justice of Hispanic descent}


Although the Treasury will announce details of the November refunding on Monday, the funding will be delayed 
if Congress and President Bush fail to increase the Treasury's borrowing capacity.

OLLIE:
(14) the Treasury                   will announce       details of the November refunding
(15) Congress and President Bush    fail to increase    the Treasury's borrowing capacity

ClausIE:
(16) the Treasury    will announce      details of the November refunding on Monday
(17) the Treasury    will announce      details of the November refunding
(18) the funding     will be delayed    if Congress and President Bush fail to increase the Treasury 's [...]
(19) the funding     will be delayed    if Congress and President Bush fail to increase the Treasury 's [...]
                                   Although the Treasury will announce details of the November [...]
(20) Congress and President Bush    fail    to increase the Treasury 's borrowing capacity
(21) the Treasury    has    borrowing capacity

\textbf{Graphene:}
\textbf{(22) #1    0    the Treasury    will announce    details of the November refunding}
\textbf{("a)     S:TEMPORAL    on Monday}
\textbf{("b)     L:CONTRAST     #2}
\textbf{(23) #2    0    the funding    will be delayed}
\textbf{("a)     L:CONTRAST     #1}
\textbf{("b)     L:CONDITION    #3}
\textbf{("c)     L:CONDITION    #4}
\textbf{(24) #3    1    Congress    fail    to increase the Treasury 's borrowing capacity}
\textbf{(25) #4    1    president Bush    fail    to increase the Treasury 's borrowing capacity}
    \end{BVerbatim}
  \caption{Comparison of the output generated by different state-of-the-art Open IE systems. Contextual arguments of Graphene are differentiated between simple textual arguments (S:) or arguments that link to other propositions (L:). Both contextual types are semantically classified by rhetorical relations.}
  \label{ComparativeAnalysis}
\end{figure}

First, relations often span over long nested structures or are presented in a non-canonical form that cannot be easily captured by a small set of extraction patterns. Therefore, such relations are commonly missed by state-of-the-art approaches. Consider for example the first sentence in Figure \ref{ComparativeAnalysis} which asserts that \texttuple{Sonia Sotomayor}{became}{the first Supreme Court Justice of Hispanic descent}. This information is encoded in a complex participial construction that is omitted by both reference Open IE systems, \textsc{Ollie} \cite{Mausam12} and ClausIE \cite{DelCorro13}.

Second, current Open IE systems tend to extract propositions with long argument phrases that can be further decomposed into meaningful propositions, with each of them representing a separate fact. Overly specific constituents that mix multiple - potentially semantically unrelated - propositions are difficult to handle for downstream applications, such as question answering (QA) or textual entailment tasks. Instead, such approaches benefit from extractions that are as compact as possible. This phenomenon can be witnessed particularly well in the extractions generated by ClausIE ((8-10) and (16-21)), whose argument phrases frequently combine several semantically independent statements. For instance, the argument in proposition (8) contains three unrelated facts, namely a direct object $\langle$\textit{Sonia Sotomayor}$\rangle$, a temporal expression $\langle$\textit{on May 26 2009}$\rangle$ and a phrasal description $\langle$\textit{to replace David Souter}$\rangle$ specifying the purpose of the assertion on which it depends.

Third, they lack the expressiveness needed to properly represent complex assertions, resulting in  incomplete, uninformative or incoherent propositions that have no meaningful interpretation or miss critical information asserted in the input sentence. Most state-of-the-art systems focus on extracting binary relationships without preserving the semantic connection between the individual propositions. This can be seen in the second example in Figure \ref{ComparativeAnalysis}. Here, the proposition \texttuple{Congress and President Bush}{fail to increase}{the Treasury's borrowing capacity} is not asserted by the input sentence, but rather represents the precondition for the predication that \texttuple{the funding}{will be delayed}{$\emptyset$}. However, both state-of-the-art Open IE systems fail to accurately reflect this contextual information in the extracted relational tuples. While \textsc{Ollie} completely ignores above-mentioned inter-proposition relationship in its extractions, ClausIE incorporates this information in its argument phrases ((18) and (19)), yet producing over-specified components that is likely to hurt the performance of downstream semantic applications.

To overcome these limitations, we developed an Open IE framework that transforms complex NL sentences into clean, compact structures that present a canonical form which facilitates the extraction of accurate, meaningful and complete propositions. The contributions of our work are two-fold. First, to remove the complexity of determining intricate predicate-argument structures with variable arity from syntactically complex input sentences, we propose a two-layered transformation process consisting of a clausal and phrasal disembedding layer. It removes clauses and phrases that convey no central information from the input and converts them into independent sentences, thereby reducing the source sentence to its main information. In that way, the input is transformed into a \textbf{novel hierarchical representation in the form of core facts and accompanying contexts}. Second, we \textbf{identify the rhetorical relations by which core sentences and their associated contexts are connected in order to preserve their semantic relationship} (see the output of our reference Open IE implementation Graphene in Figure~\ref{ComparativeAnalysis}). These two innovations enable us to enrich extracted relational tuples of the form $\langle arg_1, rel, arg_2 \rangle$ with contextual information that further specifies the tuple and to establish semantic links between them, allowing to fully reconstruct the informational content of the input.
The resulting representation of the source text
can then be used to facilitate a variety of artificial intelligence tasks, such as building QA systems or supporting semantic inferences. The idea of generating a syntactically sound representation from linguistically complex NL sentences can be easily transported to other tasks than Open IE, e.g. it might be helpful for problems such as sentiment analysis, coreference resolution or text summarization.

\section{Related Work}\label{relWork}

\paragraph{Learning-based approaches.} The line of work on Open IE begins with \textsc{TextRunner} \cite{Banko07}, a self-supervised learning approach which uses a Naive Bayes classifier to train a model of relations over examples of extraction tuples that are heuristically generated from sentences in the Penn Treebank using unlexicalized POS and NP chunk features. The system then applies the learned extractor to label each word between a candidate pair of NP arguments as part of a relation phrase or not. \textsc{WOE} \cite{WuFei10} also learns an open information extractor without direct supervision. It makes use of Wikipedia as a source of training data by bootstrapping from entries in Wikipedia infoboxes to learn extraction patterns on both POS tags (\textsc{WOE}\textit{\textsuperscript{pos}}) and dependency parses (\textsc{WOE}\textit{\textsuperscript{parse}}). By comparing their two approaches, \newcite{WuFei10} show that the use of dependency features results in an increase in precision and recall over shallow linguistic features (though, at the cost of extraction speed). \textsc{Ollie} follows the idea of bootstrap learning of patterns based on dependency parse paths. However, while \textsc{WOE} relies on Wikipedia-based bootstrapping, \textsc{Ollie} applies a set of high precision seed tuples from its predecessor system \textsc{ReVerb} to bootstrap a large training set. Moreover, \textsc{Ollie} is the first Open IE approach to identify not only verb-based relations, but also noun-mediated ones.

\paragraph{Rule-based approaches.} The second category of Open IE systems make use of hand-crafted extraction rules. This includes \textsc{ReVerb} \cite{Fader11}, a shallow extractor that applies a set of lexical and syntactic constraints that are expressed in terms of POS-based regular expressions. In that way, the amount of incoherent, uninformative and overspecified relation phrases is reduced. While previously mentioned Open IE systems focus on the extraction of binary relations, \textsc{KrakeN} \cite{Akbik12} is the first approach to be specifically built for capturing complete facts from sentences by gathering the full set of arguments for each relation phrase within a sentence, thus producing tuples of arbitrary arity. The identification of relation phrases and their corresponding arguments is based on hand-written extraction rules over typed dependency information. \textsc{Exemplar} \cite{Mesqu13} applies a similar approach for extracting n-ary relations, as it uses hand-crafted patterns based on dependency parse trees to detect a relation trigger and the arguments connected to it.

\paragraph{Clause-based approaches.} Aiming to improve the accuracy of Open IE approaches, more recent work is based on the idea of incorporating a sentence re-structuring stage whose goal is to transform the original sentence into a set of independent clauses that are easy to segment into Open IE tuples. An example of such a paraphrase-based Open IE approach is ClausIE, which exploits linguistic knowledge about the grammar of the English language to map the dependency relations of an input sentence to clause constituents. In that way, a set of coherent clauses presenting a simple linguistic structure is derived from the input. Then, one or more predicate-argument extractions are generated for each clause. In the same vein, \newcite{Schmid14} propose a strategy to break down structurally complex sentences into simpler ones by decomposing the original sentence into its basic building blocks via chunking. The dependencies of each two chunks are then determined using dependency parsing or a Naive Bayes classifier. Depending on their relationships, chunks are combined into simplified sentences, upon which the task of relation extraction is carried out. \newcite{Angeli15} present Stanford Open IE, an approach in which a classifier is learned for splitting a sentence into a set of logically entailed shorter utterances which are then maximally shortened by running natural logic inference over them. In the end, a small set of hand-crafted patterns are used to extract a predicate-argument triple from each utterance.

On the basis of such re-arrangement strategies for decomposing a complex input sentence into a set of self-contained clauses that present a linguistic structure that is easier to process for Open IE systems, we have developed an Open IE pipeline which will be presented in the following section.

\section{Proposed Open IE Approach}

We propose a method for facilitating the task of Open IE on sentences that present a complex linguistic structure. It is based on the idea of disembedding clausal and phrasal constituents out of a source sentence. In doing so, our approach identifies and retains the semantic connections between the individual components, thus generating a novel lightweight semantic Open IE representation.

\begin{figure*}[ht]
\centering

\begin{minipage}{0.17\textwidth}
\centering
Input-Document\\
\vspace{0.3cm}
\begin{tikzpicture}[scale=0.55, every node/.style={align=center, transform shape}]
\node[rectangle, solid, draw=black, text width=0.85*\columnwidth, rounded corners=5pt]{
[...] Although the Treasury will announce details of the November refunding on Monday, the funding will be delayed if Congress and President Bush fail to increase the Treasury's borrowing capacity. [...]
};
\end{tikzpicture}
\end{minipage}%
\hspace{-0.5cm}
$\rightarrow$
\begin{minipage}{0.35\textwidth}
\centering
Transformation Stage\\
\vspace{0.3cm}
\begin{tikzpicture}[scale=0.55, level distance=2cm, sibling distance=0cm, every tree node/.style={align=center, transform shape}]
\Tree [
.\node[style={draw,rectangle}] {DOCUMENT-ROOT}; 
  \edge node[midway, right] {core}; [
      .\node [style={draw,rectangle}] {Coordination\\\textit{Contrast}};
            \edge node[midway, left] {core}; [.\node(a)[label=below:TEMPORAL(on Monday)]{The Treasury will\\announce details of\\the November refunding.};]
            \edge node[midway, right] {core}; [
              .\node [style={draw,rectangle}] {Subordination\\\textit{Condition}};
                \edge node[midway, left] {core}; [.\node(b){The funding\\will be delayed.};]
                \edge node[midway, right] {context}; [
                  .\node [style={draw,rectangle}] {Coordination\\\textit{List}};
                    \edge node[midway, left] {core}; [.\node(c){Congress fail to\\increase the Treasury's\\borrowing capacity.};]
                    \edge node[midway, right] {core}; [.\node(d){President Bush fail to\\increase the Treasury's\\borrowing capacity.};]
                ]
            ]
        ]
    ]
]
\end{tikzpicture}
\end{minipage}%
$\rightarrow$
\hspace{0.5cm}
\begin{minipage}{0.35\textwidth}
\centering
Relation Extraction\\
\vspace{0.3cm}
\begin{tikzpicture}[scale=0.55, every node/.style={align=center, transform shape}]
    \node(A)[text width=0.6*\linewidth, label=below:TEMPORAL(on Monday)]{The Treasury will announce details [...]};
    \node(B)[below of=A, yshift=-1cm, text width=0.6*\linewidth]{The funding will be delayed.};
    \node(C)[below of=B, yshift=-0.5cm, text width=0.6*\linewidth]{Congress fail to increase [...]};
    \node(D)[below of=C, yshift=-0.5cm, text width=0.6*\linewidth]{President Bush fail to increase [...]};

    \node(a)[right=1 of A, text width=0.6*\linewidth,  label=below:TEMPORAL(on Monday)]{\texttuple{The Treasury}{will announce}{details [...]}};
    \node(b)[right=1 of B, text width=0.6*\linewidth]{\texttuple{The funding}{will be delayed}{}};
    \node(c)[right=1 of C, text width=0.6*\linewidth]{\texttuple{Congress}{fail}{to increase [...]}};
    \node(d)[right=1 of D, text width=0.6*\linewidth]{\texttuple{President Bush}{fail}{to increase [...]}};

    \draw[solid, <->] (A)..controls +(west:3) and +(west:3)..([yshift=10]B) node [left, midway] () {Contrast};
    \draw[solid, ->] (B)..controls +(west:3) and +(west:3)..([yshift=5]C) node [left, midway] () {Condition};
    \draw[solid, ->] ([yshift=-10]B)..controls +(west:3) and +(west:3)..([yshift=-5]D) node [left, midway] () {Condition};
    \draw[solid, <->] ([yshift=-5]C)..controls +(west:2.5) and +(west:2.5)..([yshift=5]D) node [right, midway] () {List};

    \draw[solid, ->] (A) to (a);
    \draw[solid, ->] (B) to (b);
    \draw[solid, ->] (C) to (c);
    \draw[solid, ->] (D) to (d);

\end{tikzpicture}\\
\end{minipage}

\caption{Extraction workflow for an example sentence.}
\label{fig:workflow}
\end{figure*}
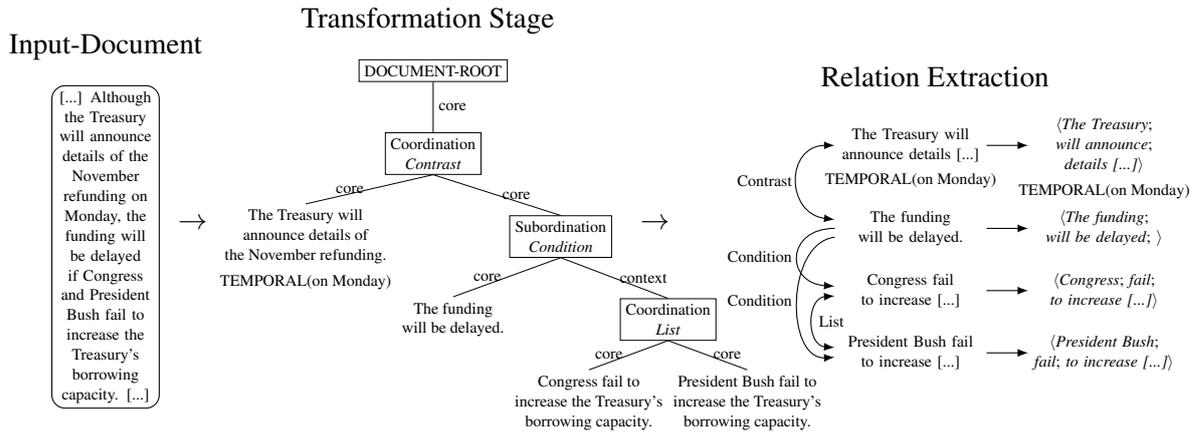

We build upon the concept presented in \newcite{Niklaus2016}, who distinguish between core and contextual information in the context of sentence simplification. This is done by disembedding and transforming supplementary material expressed in phrases (e.g. spatial or temporal information) into stand-alone context sentences, thus reducing the input sentences to their key information (core sentences). In our work, we now port this idea to a broader scope by targeting clausal disembedding techniques for complex, nested structures. Furthermore, by converting the whole simplification process into a recursive process, we are able to generate a hierarchical representation of syntactically simplified core and contextual sentences (see center image of Figure~\ref{fig:workflow}). Moreover, this allows us to detect both local and long-range rhetorical dependencies that hold between nested structures, similar to Rhetorical Structure Theory (RST) \cite{mann1988rhetorical}. As a consequence, when carrying out the task of Open IE on the compressed core sentences, the complexity of determining intricate predicate-argument structures is removed and the extracted propositions can be semantically linked and enriched with the disembedded contextual information.
Our proposed framework uses a two-layered transformation stage
for recursive sentence simplification that is followed by a final relation extraction stage. It takes a text document as an input and returns a set of hierarchically ordered and semantically interconnected relational tuples (see the output of Graphene in Figure~\ref{ComparativeAnalysis}). The workflow of our approach is displayed in Figure~\ref{fig:workflow}.

\subsection{Transformation Stage}
The core component of our work is the two-layered transformation stage where sentences that present a complex linguistic form are recursively transformed into simpler, compact, syntactically sound sentences with accompanying contextual information.

\subsubsection{Concept of the Discourse Tree Creation}
During the transformation process, we aim to identify intra-sentential semantic relationships which can later be used to restore semantic relations between extracted propositions.
To capture these semantic relations, we employ a concept of RST, in which a document is represented as a hierarchy of consecutive, non-overlapping text spans (so-called Elementary Discourse Units (EDUs)) that are connected by rhetorical relations such as \textit{Condition}, \textit{Enablement} or \textit{Background}. In addition, \newcite{mann1988rhetorical} introduced the notion of \textit{nuclearity}, where text spans that are connected by rhetorical relations  are classified as either \textit{nucleus}, when embodying the central part of information, or \textit{satellite}, whose role is to further specify the nucleus.
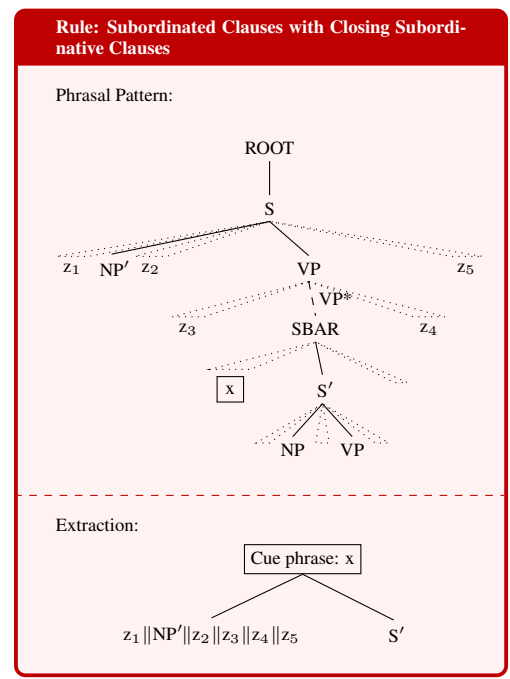
\begin{wrapfigure}{r}{6.5cm}
\centering
\scriptsize
\begin{definitionbox}{Rule: \ruleSubordinationPost}
Phrasal Pattern:
\begin{center}
\begin{tikzpicture}[level distance=0.8cm, every tree node/.style={align=center}]
\Tree [.ROOT
        [.S
          \edge[roof, dotted]; {z_1}
          NP\1
          \edge[roof, dotted]; {z_2}
          [.VP
            \edge[roof, dotted]; {z_3}
            \edge[dashed] node[midway, right] {VP*}; [.SBAR 
              \edge[roof, dotted]; {\framebox{x}}
              [.S\1 \edge[roof, dotted]; {} NP \edge[roof, dotted]; {} VP \edge[roof, dotted]; {} ]
              \edge[roof, dotted]; {}
            ]
            \edge[roof, dotted]; {z_4}
          ]
          \edge[roof, dotted]; {z_5}
        ] 
  ]
\end{tikzpicture}
\end{center}
\tcbline
Extraction:
\begin{center}
\begin{tikzpicture}[level distance=1cm, sibling distance=1cm, every tree node/.style={align=center}]
\Tree [.\node[style={draw,rectangle}] {Cue phrase: x}; 
  \edge node[midway, left] {}; {z_1 \conc NP\1 \conc z_2 \conc z_3 \conc z_4 \conc z_5}
  \edge node[midway, right] {}; {S\1}
]
\end{tikzpicture}
\end{center}
\end{definitionbox}

\caption{Rule for \MakeLowercase{\ruleSubordinationPost} ("\conc" denotes concatenation).}
\label{fig:subordination_post_definition}
\end{wrapfigure}
We adapt the RST framework for our task of clausal sentence simplification, where the goal is to split up complex multi-clause sentences in a recursive fashion in order to obtain a hierarchical structure of the input similar to the diagrams used in RST. As opposed to RST, the resulting simplified sentences, which are represented as leaf nodes in the tree, cannot be specified prior to the transformation process, since they are not uniform and might require transformations (e.g. paraphrasing) for specific syntactic environments in which they occur. Therefore, the transformation process is carried out in a top-down fashion, starting with the input document and using a set of syntactic rule patterns that define how to split up, transform and recurse on complex syntactic sentence patterns (see Section~\ref{sec:transformationRules}). Each split will then create two or more simplified sentences that are connected with information about their constituency type (\textit{coordinate} or \textit{subordinate}) and identified rhetorical relation. The constituency type infers the concept of nuclearity, where coordinate sentences (which we will call \textit{core} sentences) represent nucleus spans and subordinate sentences (\textit{context} sentences) represent satellite spans. In this way, we obtain a hierarchical tree representation of the input document similar to RST. We will denote the resulting tree as a \textit{discourse tree}. 


Note that our approach differs from RST in the following ways: 1) connected sentences do not have to be non-overlapping text spans;
2) our approach works in a top-down fashion where the final simplified sentences result from the transformation process, whereas in RST, the smallest elements of discourse (EDUs) are identified in advance.

\subsubsection{Algorithm}

\begin{algorithm}
\small
\caption{Transformation Stage}
\label{alg:transformation}
\begin{algorithmic}[1]
    \Require{the document-string $str$}
    \Ensure{the discourse tree $discourse\_tree$}
    \Statex
    \Function{Transformation}{$str$}
        \State $discourse\_tree \gets$ initialize as a document root node with leaf-nodes for each sentence in $str$.
        \State \Call{TraverseDiscourseTree}{$discourse\_tree$}
        \State $discourse\_tree \gets$ further disembed contextual phrases (e.g. $temporal$, $spatial$) out of sentence leaves using the Sentence Simplification system proposed by \newcite{Niklaus2016}. 
        \State \Return $discourse\_tree$
    \EndFunction
    \Statex
    \Procedure{TraverseDiscourseTree}{$tree$}
      \LineComment{Process leaves from left to right}
      \For {$leaf$ in $tree.leaves$}
      	\State $match \gets$ \texttt{False}
        \LineComment{Check clausal transformation rules in fixed order}
        \For {$rule$ in $TRANSFORM\_RULES$}
          \If{$rule.pattern.matches(leaf.parse\_tree)$}
            \State $match \gets$ \texttt{True}
            \State \textbf{Break}
          \EndIf
        \EndFor
        \If{$match ==$ \texttt{True}}
          \LineComment{Disembedding / Simplification}
          \State $simplified\_sentences \gets rule.disembed(leaf.parse\_tree)$
          \State $new\_leaves \gets$ convert $simplified\_sentences$ into leaf nodes. Nodes for subordinate sentences are labelled as \texttt{context}, else \texttt{core}
          \State $new\_node \gets$ create new parent node for $new\_leaves$. If a context sentence is enclosed, the constituency is \texttt{subordinate}, else \texttt{coordinate}
          \LineComment{Rhetorical relation identification}
          \State $cue\_phrase \gets rule.extract\_cue(leaf.parse\_tree)$ 
          \State $new\_node.relation \gets$ match $cue\_phrase$ against a list of rhetorical cue words that are assigned to their most likely triggered rhetorical relation. Different cue word lists are used for different syntactic environments.
          \LineComment{Recursion}
          \State $tree.replace(leaf, new\_node)$ 
          \State \Call{TraverseDiscourseTree}{$new\_node$}
        \EndIf
      \EndFor

    \EndProcedure
\end{algorithmic}
\end{algorithm}

The transformation algorithm (see Algorithm \ref{alg:transformation}) takes as an input the document text as a string and outputs its discourse tree.
In the initialization step (2), the discourse tree is set up in the form of a single, unclassified coordination node that represents the root of the document's discourse tree. It contains edges to every sentence that has been tokenized out of the document string, which are treated as coordinated leaf nodes. In this way, all sentences are considered as core sentences, thus being equally relevant.

\begin{wrapfigure}{r}{8.5cm}
\centering
\scriptsize
\begin{examplebox}{Example: \ruleSubordinationPost}
Sentence: \inlinetext{The Great Red Spot may have been observed in 1664 by Robert Hooke, although this is disputed.}
\tcbline
Matched Pattern:
\begin{center}
\begin{tikzpicture}[scale=0.75, every tree node/.style={align=center}, level distance=0.7cm]
\Tree [.ROOT
        [.S
          [.NP \edge[roof]; {The Great\\Red Spot} ]
          [.VP
            [.MD may ]
            [.VP
              [.VB have ]
              [.VP
                [.VBN been ]
                [.VP
                  [.VBN observed ]
                  [.PP \edge[roof]; {by Robert\\Hooke} ]
                  [., , ]
                  [.SBAR
                    [.IN \framebox{although} ]
                    [.S
                      [.NP [.DT this ] ]
                      [.VP \edge[roof]; {is\\disputed} ] ] ] ] ] ] ]
          [.. . ] ] ]
\end{tikzpicture}
\end{center}
\tcbline
Extraction:
\begin{center}
\begin{tikzpicture}[level distance=1cm, sibling distance=0.5cm, every tree node/.style={align=center}]
\Tree [.\node[style={draw,rectangle}] {\inlinetext{although} $\rightarrow$ \textit{Contrast}}; 
  \edge node[midway, left] {core}; {The Great Red Spot may have been\\observed in 1664 by Robert Hooke.}
  \edge node[midway, right] {core}; {This is disputed.}
]
\end{tikzpicture}
\end{center}
\end{examplebox}

\caption{Example for \MakeLowercase{\ruleSubordinationPost}.}
\label{fig:subordination_post_example}
\end{wrapfigure}



After initialization, the discourse tree is traversed and split up recursively in a top-down approach (3, 7).
The tree traversal is done by processing sentence leaves in depth-first order, starting from the root node. For every leaf (9), we check if its phrasal parse structure matches one of the fixed ordered set of 16 hand-crafted syntactic rule patterns (see Section \ref{sec:transformationRules}) that determine how a sentence is split up into two or more simplified sentence leaves (13). The first matching rule will be used, generating two or more simplified sentences (20) that will again be processed in this way.
In addition to generating simplified sentences, each rule also determines the constituency type of the newly created node (22) and returns a cue phrase which is used as a lexical feature for classifying the type of rhetorical relation connecting the split sentences (24, 25). 
The algorithm terminates when no more rule matches the set of generated sentences.
Finally, we use the sentence simplification system of \newcite{Niklaus2016} to further disembed contextual phrases out of the simplified sentences from the discourse tree (4).
The complete source code of our framework can be found online\footnote{\url{https://github.com/Lambda-3/Graphene}}.



\subsubsection{Transformation Rules}
\label{sec:transformationRules}

The set of hand-crafted transformation rules used in the simplification process are based on syntactic and lexical features that can be obtained from a sentence's phrase structure, which we generated with the help of Stanford's pre-trained lexicalized parser \cite{Socher2013}. These rules make use of regular expressions over the parse trees encoded in the form of Tregex patterns \cite{Levy2006}. 
They were heuristically determined in a rule engineering process whose main goal was to provide a best-effort set of rules, targeting the challenge of being applied in a recursive fashion and to overcome biased or erroneous parse trees. During our experiments, we developed a fixed execution order of rules which achieved the highest F\textsubscript{1}-score in the evaluation setting.

Each rule pattern accepts a sentence's phrasal parse tree as an input and encodes a certain parse tree pattern that, in case of a match, will extract textual parts out of the tree that are used to produce the following information:

\begin{description}
    \item[Simplified sentences] The simplified sentences are generated by combining extracted parts from the parse tree. Note that some of the extracted parts such as phrases also require to be arranged and complemented in a way so that they form grammatically correct sentences (\textit{paraphrasing}), aiming for a canonical subject-predicate-object structure.

    \item[Constituency] Our notion of constituency depicts the contextual hierarchy between the simplified sentences. If all sentences can be considered to be equally important, we will call it a \textit{coordination}. In this case, all generated sentences are labeled as \textit{core} sentences. Otherwise, i.e. if one sentence provides background information or further specifies another sentence, we use the term \textit{subordinate} and label the sentence as \textit{context} sentence.
   In most of the cases, our framework uses the syntactic information of coordinated and subordinated clauses to infer the same type of (semantic) constituency with respect to the rhetorical relations. An example where this does not apply are rules for identifying attributions. In these cases, we consider an actual statement (e.g. \inlinetext{Economic activity continued to increase.}) to be more important than the fact that this was stated by some entity (e.g. \inlinetext{This was what the Federal Reserve noted.}).

    \item[Classified rhetorical relation] Both syntactic and lexical features are used during the transformation process to identify and classify rhetorical relations that hold between the simplified sentences.
    Syntactic features are manifested in the phrasal composition of a sentence's phrasal parse tree, whereas lexical features are extracted from the parse tree as so-called \textit{cue phrases}. Cue phrases (often denoted as \textit{discourse markers}) represent a sequence of words that are likely to include rhetorical cue words (e.g. \inlinetext{because}, \inlinetext{after that} or \inlinetext{in order to}) indicating a certain rhetorical relation. The way these cue phrases are extracted from a sentence's parse tree is specific to each rule pattern. For determining potential cue words and their positions in specific syntactic environments, we use the work of \newcite{knott1994using}. 
    The extracted cue phrases are then used to infer the kind of rhetorical relation. For this task, we use a predefined list of rhetorical cue words which are assigned to the rhetorical relation that they most likely trigger. The mapping of cue words to rhetorical relations was adapted from the work of \newcite{Taboada13}. If one of these cue words is present in the cue phrase, the corresponding rhetorical relation is set between the newly constructed sentences. 
\end{description}


An example for a transformation rule, targeting subordinate clausal constructions, is illustrated in Figure ~\ref{fig:subordination_post_definition}. Figure~\ref{fig:subordination_post_example} shows its application to an example sentence. The full rule set can be found in the supplementary material online\footnote{\url{https://github.com/Lambda-3/Graphene/tree/master/wiki/supplementary}}. Here, a detailed example that demonstrates the process of turning a complex NL sentence into a discourse tree is provided, too.

\subsection{Relation Extraction}

\begin{wrapfigure}{r}{7cm}
\centering
\begin{minipage}{.15\textwidth}
\centering
\vspace{1.6cm}
$rel \gets$ $z_3$ \conc $z_4$\\
$arg_1 \gets$ NP'\\
$arg_2 \gets$ x \conc $z_5$
\end{minipage}%
\hspace{-1.5cm}
\begin{minipage}{.32\textwidth}
\centering
\begin{tikzpicture}[scale=0.76, level distance=1.2cm, every tree node/.style={align=center, transform shape}]
\Tree [.{topmost S\1}
          \edge[roof, dotted]; {z_1}
          NP\1
          \edge[roof, dotted]; {z_2}
          [.VP
            \edge[roof, dotted]; {z_3}
            \edge[dashed] node[midway, right] {VP*};
            [
              \edge[roof, dotted]; {z_4}
              {NP $\vert$ PP $\vert$\\S $\vert$ SBAR\\x}
              \edge[roof, dotted]; {z_5}
            ]
            \edge[roof, dotted]; {z_6}
          ]
          \edge[roof, dotted]; {z_7}
        ] 
\end{tikzpicture}
\end{minipage}%

\caption{Graphene's default relation extraction pattern ("\conc" denotes concatenation).}
\label{fig:headRelationExtractor}
\end{wrapfigure}
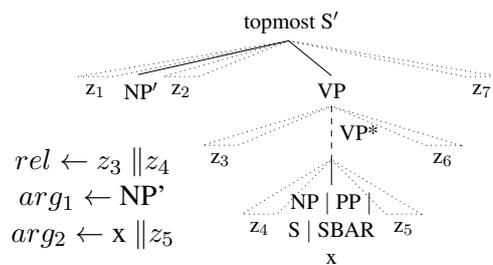
The last step in our framework is realized by the actual relation extraction task. To do so, each of the previously generated leaf sentences is given as an input to any Open IE system that is able to extract one (or more) binary propositions from a given sentence. In order to map the identified rhetorical and contextual relationships from the discourse tree to the extracted propositions, we need to determine which of the extracted propositions embodies the main statement of the corresponding sentence. We use the following heuristic to decide whether an extracted proposition represents the main statement of the input sentence: 1) the head verb of the input sentence is contained in the relational phrase $rel$ of the proposition \tuple{$arg_1$}{$rel$}{$arg_2$} (e.g. \texttuple{Employees}{are \ul{nominated}}{for the program}); or 2) the head verb of the input sentence equals the object argument $arg_2$ of the proposition (e.g. \texttuple{Employees}{are}{\ul{nominated}}). We will denote any proposition that matches those criteria as a \textit{representative} of the corresponding sentence. Once determined, the identified rhetorical and contextual relations that hold between two sentences from the discourse tree can be directly transferred as (semantically typed) additional arguments to the corresponding representatives, thus generating n-ary relations.

 
In this way, extractions from any Open IE system that uses our framework can be enriched with semantic information. An example of the semantically enriched output produced by the Open IE system \textsc{OLLIE} when using our framework as a preprocessing step is shown in Figure~\ref{ComparativeAnalysisSystems}.
In our baseline implementation Graphene, we use one syntactic rule pattern that operates on the sentence's phrase structure and identifies simple subject-predicate-object structures. The default pattern, which extracts one tuple for each sentence, is shown in Figure~\ref{fig:headRelationExtractor}.

\begin{figure}[!ht]
  \centering
    \scriptsize
    \begin{BVerbatim}[commandchars=\\\{\},codes={\catcode`$=3\catcode`_=8}]
He nominated Sonia Sotomayor on May 26, 2009 to replace David Souter; she was confirmed on August 6, 2009, 
becoming the first Supreme Court Justice of Hispanic descent. 

OLLIE (alone):
(1) she                was confirmed on                   August 6, 2009
(2) He                 nominated Sonia Sotomayor on       May 26
(3) He                 nominated Sonia Sotomayor          2009
(4) He                 nominated 2009 on                  May 26
(5) Sonia Sotomayor    be nominated 2009 on               May 26
(6) He                 nominated 2009                     Sonia Sotomayor
(7) 2009               be nominated Sonia Sotomayor on    May 26   

Ollie (using framework):
(8)  #1    0    he    nominated    Sonia Sotomayor
("a)     S:PURPOSE     to replace David Souter .
("b)     S:TEMPORAL    on May 26 , 2009 .
(9)  #2    0    she    was becoming    the first Supreme Court Justice of Hispanic descent
    \end{BVerbatim}
  \caption{Comparison of the output of \textsc{Ollie} alone and with the framework.}
  \label{ComparativeAnalysisSystems}
\end{figure}

\section{Evaluation}

In order to evaluate the performance of our Open IE reference implementation Graphene, we conduct an automatic evaluation using the Open IE benchmark framework proposed in \newcite{Stanovsky2016EMNLP}, which is based on a QA-Semantic Role Labeling (SRL) corpus with more than 10,000 extractions over 3,200 sentences from Wikipedia and the Wall Street Journal\footnote{available under \url{https://github.com/gabrielStanovsky/oie-benchmark}}. This benchmark allows us to compare our system with a variety of current Open IE approaches in recall and precision. Moreover, we investigate whether our two-layered transformation process of clausal and phrasal disembedding improves the performance of state-of-the-art Open IE systems when applied as a preprocessing step.

\subsection{Experimental Setup}

To conduct an automatic comparative evaluation, we integrate Graphene into \newcite{Stanovsky2016EMNLP}'s Open IE benchmark framework, which was created from a QA-SRL dataset where every verbal predicate was considered as constituting an own extraction. Since Graphene's structured output is not designed to yield extractions for every occurrence of a verbal predicate, we use an alternative relation extraction implementation  which is able to produce more than one extraction from a simplified core sentence.
To match extractions to reference propositions from the gold standard, we use the method described in \newcite{Stanovsky2016EMNLP}, following \newcite{he2015question} by matching an extraction with a gold proposition if both agree on the grammatical head of both the relational phrase and its arguments. Since n-ary relational tuples with possibly more than two arguments have to be compared, we assign a positive match if the relational phrase and at least two of their arguments match.
Since the gold standard is composed of n-ary relations, we add all contexts ($S$) of an extraction as additional arguments besides $arg_1$ and $arg_2$. 
We assess the performance of our system together with the state-of-the-art systems ClausIE, \textsc{Ollie}, OpenIE-4 \cite{Mausam16}, PropS \cite{StanovskyFDG16}, \textsc{ReVerb} and Stanford Open IE.

After evaluating Graphene as a reference implementation of the proposed approach, we investigate how clausal and phrasal disembedding applied as a preprocessing step affects the performance of other Open IE systems. Therefore, we compare the performance of ClausIE, \textsc{Ollie}, OpenIE-4, \textsc{ReVerb} and Stanford Open IE on the raw input data with their performance when operating inside of our framework where they act as different relation extraction implementations that take the simplified sentences of the transformation stage as an input.

\subsection{Results and Discussion}

\begin{figure}[ht]
\begin{minipage}[ht]{0.5\textwidth}
	\centering
	\includegraphics[width=7.5cm]{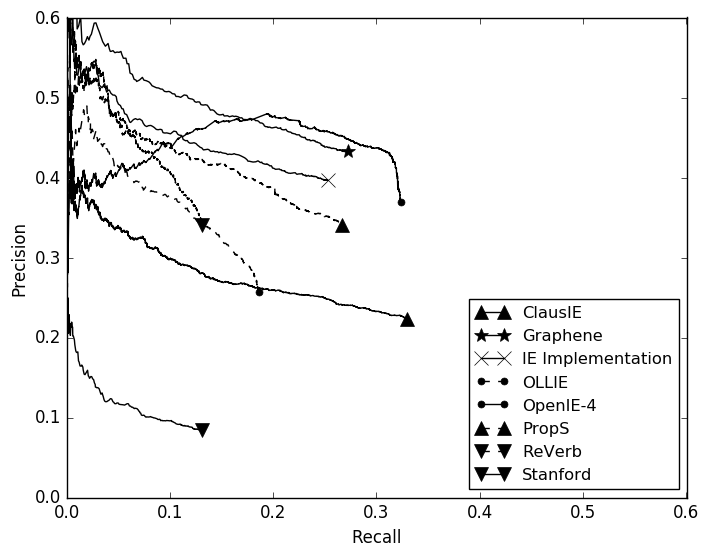}  
  \caption{Performance of Graphene.}
  \label{fig:OIEBenchmark}
\end{minipage}
\hfill
\begin{minipage}[ht]{{0.5\textwidth}}
	\centering
	\includegraphics[width=7.5cm]{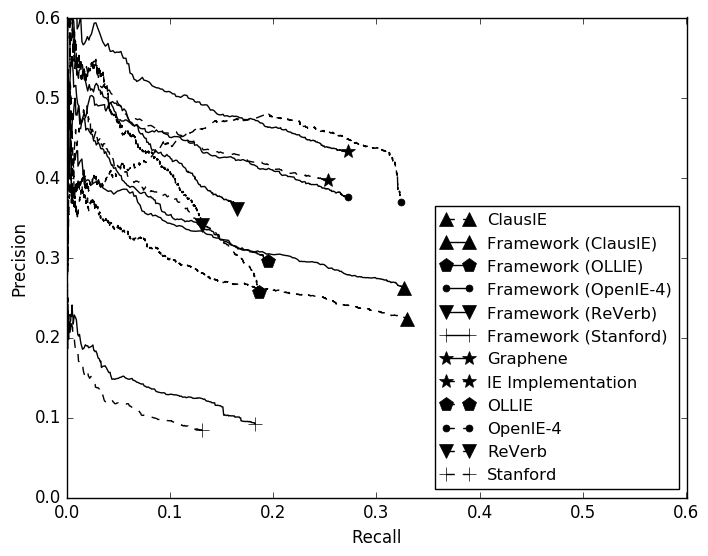}  
  \caption{Improvements of state-of-the-art systems when operating as RE component of our framework.}
  \label{fig:OIEBenchmarkSystems}
\end{minipage}
\end{figure}

Figure~\ref{fig:OIEBenchmark} shows the precision-recall curve for each system within the Open IE benchmark evaluation framework.
In our experiments, with a score of 50.1\% in average precision, our reference implementation Graphene achieves the best performance of all the systems in extracting accurate relational tuples, followed by OpenIE-4 (44.6\%) and PropS (42.4\%).
Considering recall, Graphene (27.2\%) is able to compete with other high-precision systems, such as PropS (26.7\%). However, it does not reach the recall rate of ClausIE (33.0\%) or OpenIE-4 (32.5\%). This lack in recall was expected, since Graphene was not designed to create fine-grained relations for every possible verb, as opposed to the gold standard, but rather determines the main relations with attached (contextual) arguments that often contain some of these verbal expressions. In an in-depth analysis of the output, we proved that only 33.72\% of unmatched gold extractions were caused by a wrong argument assignment of Graphene. In the majority of cases (66.28\%), Graphene did not extract propositions with a matching relational phrase. We further calculated that in 56.80\% of such cases, the relational phrase of a missed gold standard extraction was actually contained inside an argument, e.g.:

\inlinetext{Japan may be a tough market for outsiders to penetrate, and the U.S. is hopelessly behind Japan in certain technologies.}
\begin{itemize}
	\item Unmatched Gold-Extraction: \texttuple{outsiders}{may \ul{penetrate}}{Japan}
	\item Graphene-Extractions:
    \begin{enumerate}
        \item \texttuple{Japan}{may be}{a tough market for outsiders to \ul{penetrate}}
		\item \texttuple{the U.S.}{is hopelessly behind Japan}{in certain technologies}
    \end{enumerate}
\end{itemize}

This number even grows to 67.89\% when considering the lemmatized form of the head:

\inlinetext{The seeds of ``Raphanus sativus'' can be pressed to extract radish seed oil.}
\begin{itemize}
	\item Unmatched Gold-Extraction: \texttuple{radish seed oil}{can be \ul{extracted}}{from The seeds}
	\item Graphene-Extractions:
    \begin{enumerate}
    	\item \texttuple{the seeds of ``Raphanus sativus''}{can be pressed}{to \ul{extract} radish seed oil}
    \end{enumerate}
\end{itemize}

Another 5.22\% of unmatched relational phrases would have been recognized by Graphene if they were compared based on their lemmatized head.


Regarding the overall Area Under the Curve (AUC) score, OpenIE-4 is the best performing system with a score of 14.5\%, closely followed by Graphene (13.6\%), PropS (11.3\%) and ClausIE (9.3\%).

The precision-recall curve in Figure~\ref{fig:OIEBenchmarkSystems} shows that when using clausal and phrasal disembedding provided by our framework, all tested systems except for OpenIE-4 (-18\%) gain in AUC.
The highest improvement in AUC was achieved by Stanford Open IE, yielding a 63\% AUC increase over the output of Stanford Open IE as a standalone system. AUC scores of \textsc{Ollie} and \textsc{ReVerb} improve by 34\% and 22\%. While \textsc{Ollie} and \textsc{ReVerb} primarily profit from a boost in recall (+4\%, +26\%), ClausIE mainly enhances precision (+16\%). Furthermore, the performance of our reference relation extraction implementation increases both in precision (+10\%) and recall (+8\%) when applied as the relation extraction component inside of our system Graphene (+19\% AUC).

\section{Conclusion}

In this paper, we introduce a novel Open IE approach that presents an innovative two-layered hierarchical representation of syntactically simplified sentences in the form of core facts and accompanying contexts that are semantically linked by rhetorical relations. In that way, the semantic connection of the individual components is preserved, thus allowing to fully reconstruct the informational content of the input.


The results of a comparative analysis conducted on a large-scale benchmark framework showed that with a score of 50.1\%, our baseline system Graphene achieves the best average precision, while also providing a recall rate of 27.2\% that is comparable to that of other high-precision systems.
Moreover, we demonstrated that by using clausal and phrasal disembedding as a preprocessing step, the AUC score of state-of-the-art Open IE systems can be improved by up to 63\%.
In summary, we were able to show that by generating a two-layered representation of core and contextual information, the relations extracted by state-of-the-art Open IE systems can be semantically enriched without losing in precision or recall. Moreover, by using clausal and phrasal disembedding techniques, contextual information is detached from the core propositions and transformed into additional arguments, thus avoiding the problem of overspecified argument phrases and yielding a more compact structure. By enhancing the resulting syntactically sound sentence representation with rhetorical relations, semantically typed and interconnected relational tuples are created that may benefit downstream artificial intelligence tasks.
In future work, we plan to evaluate the classification of the rhetorical relations, examine to what extent other NL Processing tasks such as QA systems may benefit from the results produced by our framework and investigate the creation of big knowledge graphs for QA by performing a large-scale Wikipedia extraction.

\bibliographystyle{acl}
\bibliography{coling2018}

\end{document}